\documentclass[conference,a4paper]{IEEEtran}

\ifCLASSINFOpdf
\else
\fi
\usepackage{subfigure,cite}
\usepackage{epsfig}
\usepackage{epspdfconversion}
\usepackage{amsmath}
\usepackage{amssymb}
\usepackage{color}

\DeclareMathAlphabet{\mathpzc}{OT1}{pzc}{m}{it} 

\newcommand{\trans}[1]{{#1}^{\ensuremath{\mathsf{T}}}} 

\newcommand{\etal}{et al.\!}
\newcommand{\eg}{e.g.\!}
\newcommand{\ie}{i.e.\!}

\begin{document}
%
\title{Crowd Saliency Detection \\ via Global Similarity Structure}



%
\author{\IEEEauthorblockN{Mei Kuan Lim\IEEEauthorrefmark{1}\IEEEauthorrefmark{3},
Ven Jyn Kok\IEEEauthorrefmark{1}\IEEEauthorrefmark{3},
Chen Change Loy\IEEEauthorrefmark{2} and
Chee Seng Chan\IEEEauthorrefmark{3}}
\IEEEauthorblockA{\IEEEauthorrefmark{3}Center of Image and Signal Processing, University of Malaya, 50603 Kuala Lumpur, Malaysia\\
Email: \{imeikuan;venjyn.kok\}@siswa.um.edu.my; cs.chan@um.edu.my}
\IEEEauthorblockA{\IEEEauthorrefmark{2}The Chinese University of Hong Kong, Shatin, NT, Hong Kong\\
Email: ccloy@ie.cuhk.edu.hk}}


\maketitle

{\let\thefootnote\relax\footnotetext{* Mei Kuan Lim and Ven Jyn Kok contributed equally to this paper.}}

\begin{abstract}
It is common for CCTV operators to overlook interesting events
taking place within the crowd due to large number of people in the crowded scene (\ie~marathon, rally). Thus, there is a dire need to automate
the detection of salient crowd regions acquiring immediate attention for a more
effective and proactive surveillance. 
This paper proposes a novel framework to identify and localize salient regions in a crowd scene, by transforming low-level features extracted from crowd motion field into a global similarity structure. The global similarity structure representation allows the discovery of the intrinsic manifold of the motion dynamics, which could not be captured by the low-level representation. 
Ranking is then performed on the global similarity structure to identify a set of extrema. The proposed approach is unsupervised so learning stage is eliminated. Experimental results on public datasets demonstrates the effectiveness of exploiting such extrema in identifying salient regions in various crowd scenarios that exhibit crowding, local irregular motion, and unique motion areas such as sources and sinks. 

\end{abstract}


%
\IEEEpeerreviewmaketitle

\section{Introduction}

 

The increasing demands for security and public safety by the society has lead to an enormous growth in the deployment of CCTV in public spaces \cite{Valera05,GongLAP2011}. The recent Boston Marathon bombing, in particular, has ignited a pressing interest for automated video content analysis to assist the law enforcement in preventing such events to be happened again. The investigation surrounding the bombing was a missed opportunity to use technology to detect the abnormal behavior of the suspect, which leads to the tragedy \cite{Joshua13}. However, one must understand that at large events such as rallies and marathons, where crowds of hundreds or even thousands gather, video monitoring is a daunting task due to the large variations of crowd densities and severe occlusions. Moreover, the attention span of human has been shown to deteriorate after 20 minutes and manual monitoring task requires demanding, prolonged cognitive attention \cite{Liu2013}. Therefore, major research efforts are emerging towards developing solutions to identify interesting or salient regions, which could ultimately lead to unfavorable events, as a cue to direct the attention of the security personnel. 

The definition of interesting region in crowd has been causing much debates in the literature due to the subjective nature and complexity of the human behaviors. Some researchers consider any deviation from the ordinary observed events as anomaly, whereas others consider rare or outstanding event as interesting. 
Finding interesting regions in a given scene is generally accomplished by firstly learning an activity model of the scene, followed by using the learned model to identify the anomalies \cite{Kuettel10,Hospedales11,Zhou2012, Rodriguez11}. 
%
%
In this study, we take a different perspective to detect the interesting regions in extremely crowded scenes. In contrast to existing studies, our method alleviates the need for a learned model. In particular, we assume that the motion of individuals tend to follow the regular or dominant flow of a particular region due to the physical structure of the scene, and the social conventions of the crowd dynamics. With this assumption, we consider interesting regions as extrema in the underlying crowd motion dynamics in the scene. Detecting these extrema is accomplished in an unsupervised manner.


\begin{figure}[t]
\centering
\includegraphics[height=0.55\linewidth, width=\linewidth]{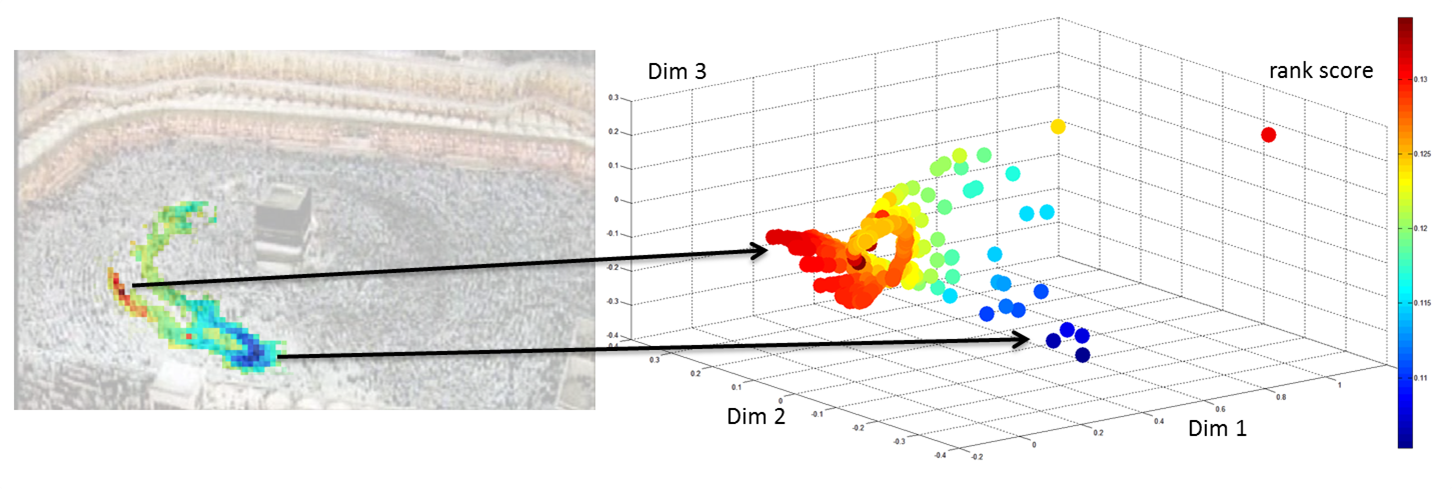} 
\caption{Three-dimensional embedding of the global similarity structure obtained using multi-dimensional scaling. The color of each point represents the ranking score, where the extrema correspond to salient regions.}
\label{fig:embedded}
\end{figure}


In contrast to existing methods \cite{ali2007lagrangian,Loy12}, which use low-level features for  crowd motion representation, we project the low-level features extracted from the motion field into a global similarity structure, which captures the pairwise similarity of the crowd motion of all pixels (or particles that are spatially distributed on the image plane).
Such a structure allows the discovery of intrinsic manifold of the motion dynamics as shown in Fig.~\ref{fig:embedded}. 
%
%
With the manifold, ranking is performed by the iterated graph Laplacian approach. The extrema of the rank scores are employed as an indicator of salient motion dynamics or unstable motion in the dense crowd scenes.
The aforementioned approach is purely unsupervised, eliminating the requirement of a learning stage as to~\cite{Kuettel10,Hospedales11,Zhou2012, Rodriguez11}. 

Experimental results on public datasets demonstrate the capability of the proposed method in detecting and localizing a broad scope of crowd salient motions caused by crowding, sources and sinks, and local irregular motion.
The crowding is defined as potential clogging or bottlenecks that are typically affected by the physical structure of the environment. For example, near junctions where the crowd density builds up and thus, preventing smooth motion amongst individuals. Sources and sinks refer to regions where individuals in a crowd enter or leave the scene. Finally, local irregular motion is triggered by flow instability of individuals or a small groups maneuvering against the dominant flow in the scene.

\section{Related Work}

Existing methods can be divided into two main approaches. The first approach analyzes crowd behaviors or activities based on the motion of individuals, where tracking of their trajectories is required \cite{makris05,wang06,Rodriguez09,Rodriguez11,nedrich10,Zhou2012,Shao2014}. Commonly, the tracking approaches keep track of each individual motion and further apply a statistical model of the trajectories to identify the semantics or geometric structures of the scene, such as the walking paths, sources and sinks. Then, the learned semantics are compared to the query trajectories to detect anomaly. While in principle individuals should be tracked from the time they enter a scene, till the time they exit the scene to infer such semantics, it is inevitable that tracking tends to fail due to occlusion, clutter background and irregular motion in the crowded scenes. Therefore, the aforementioned methods work well, up to a certain extent, in sparse crowd scenes. They tend to fail in dense crowd scenes (Fig.~\ref{fig:embedded}), where target tracking is extremely challenging. 

In order to alleviate the need to track individuals in the scene, researchers have proposed holistic approach for activity analysis and behavior understanding in the crowded scenes. Rather than computing the trajectories of individuals, this approach builds a crowd motion model using the instantaneous motions of the entire scene such as the flow field \cite{Andrade06,Hu2008}. The flow field is then fed into an hidden Markov model to learn the inherent dynamics of the motion patterns\cite{Andrade06}, or clustering methods for motion segmentation~\cite{Hu2008}.  
%
%
%
Ali~\etal~\cite{ali2007lagrangian} apply the Lagrangian particle dynamics based on the crowd flow field to estimate the stability of a particular region. Their method able to detect regions with unstable motion by discovering the abnormality in the segmented flow fields. Similarly, \cite{Mancas11} proposed another representation of the low-level features extracted from the optical flow using a multi-scale approach to identify interesting regions. Since these methods use only the direction and speed as the motion features, their scenarios are limited to those events that are occurred due to the variation in motion direction and speed only. Example of these detections include an individual moves at a faster speed than the group, or moving at the opposite direction. Their method are not able to cope with other type of saliency such as crowding, or unique motion areas such as the sources and sinks. 

Detection and localization of salient regions by using spectral analysis is proposed in \cite{Loy12}. In contrast to other methods, their method suppress dominant flows with a focus on the motion flows that deviate from the norm. 
While their method deal with unstable crowd flow, their experiments were limited to the detection of simulated instability, and not real-world public scenes. In the closest work to ours, Solmaz~\etal~\cite{solmaz12} propose a linear approximation of the dynamical system to categorize different crowd behaviors using the eigenvalues over an interval of time. Their methods show promising results in detecting and classifying five different scenarios of saliency, which includes the bottleneck, lane, arch, fountainhead and blocking. In comparison to \cite{solmaz12}, our method is more sensitive in detecting such salient regions, while having the capability of highlighting the location of the triggering event accurately.
 
In summary, the main contribution of this study is that we propose the transformation of low-level motion features into \textit{global similarity structure}. The structure allows the discovery of the intrinsic manifold of the motion dynamics in crowded scenes, which could not be captured by the low-level representation as to \cite{ali2007lagrangian,Loy12}. Moreover, contrary to the state-of-the art solutions \cite{Kuettel10,Hospedales11,Zhou2012, Rodriguez11}, the presented manifold requires (1) \textit{no tracking,} as we exploit optical flow representation, and (2) \textit{no prior information or model learning} to identify interesting/salient regions in the crowded scenes, as we employ extrema in the \textit{intrinsic manifold} of motion dynamics as an indicator of saliency.



\section{Proposed Framework}
\begin{figure}[t]
\centering
\subfigure[Input video sequence]{
    \includegraphics[height=0.25\linewidth, width=0.45\linewidth]{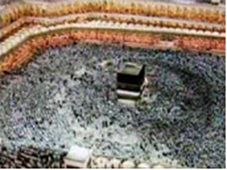} 
    \label{fig:subfigintro1f}
}
\subfigure[Motion flow estimation]{
    \includegraphics[height=0.25\linewidth, width=0.45\linewidth]{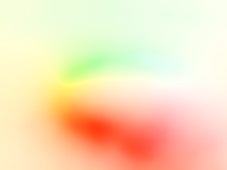} 
    \label{fig:subfigintro2f}
}
\centering
\subfigure[(Left) stability map and (Right) phase shift map reveals the global similarity structure of the scence. The width and height of the map are the number of pixels of a video frame.]{
    \includegraphics[height=0.25\linewidth, width=0.9\linewidth]{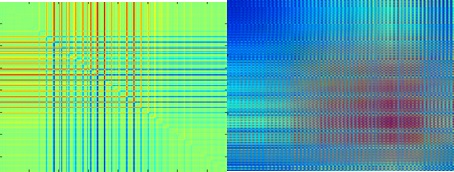} 
    \label{fig:subfigintro3f}
}
\subfigure[The ranking results, where red and blue color indicate the extrema with interesting dynamics.]{
    \includegraphics[height=0.25\linewidth, width=0.9\linewidth]{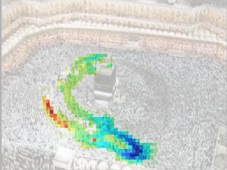} 
    \label{fig:subfigintro4f}
}
\label{fig:figuresf}
\caption{Outputs from the key steps in crowd saliency detection. Best viewed in color.}
\label{fig:framework} 
\end{figure}

The pipeline of the proposed framework is illustrated in Fig.~\ref{fig:framework}.

\subsection{Crowd Motion Field}

The proposed framework represents the crowd motion field of each frame using the optical flow. Specifically, given a crowd video sequence, the velocity field at each point, $V(p) = (u_p,v_p)$ is estimated using the dense optical flow algorithm as to \cite{liu08}, where each pixel in a given frame is considered as a point or particle\footnote{One could also consider a spatial block of pixels as a particle.}, $p=(x,y)$. Both the horizontal and vertical flow components, $u$ and $v$, of the extracted optical flow field are then accumulated, and an averaged flow, $\overline{V}$, is calculated within an interval of time, comprising $| \tau |$ frames.

\begin{equation}\label{eq:mean01}
\overline{V} = \{\overline{u},\overline{v}\} = \{\frac{1}{\tau}\sum _t^{t+\tau} u_p, \frac{1}{\tau}\sum _t^{t+\tau} v_p\}
\end{equation}
The proposed interval-based average representation is performed to obtain smooth and consistent fields, where inconsistent velocity components (noise) are often reduced if not removed during the averaging step.

\subsection{Feature Representation}

Using the crowd motion field, we extract two features to represent a broader definition of the crowd dynamics denoted as the stability and phase shift maps. These maps are the results of transformation of the low-level feature space into global similarity structure space. Next we describe the computation of each map in detail.

\subsubsection{Stability Map}

The mean optical flow field appears to be a good indicator for the dominant flow of individuals in crowd, but may not be sensitive enough to capture subtle interaction and motion flows that deviate from the norm. To this end, we apply particle advection to the mean flow field. The resulting pathlines from the advection process allows quantification of the motion dynamics, which is derived later from the separation coefficients between particles. The basic idea of particle advection is to approximate the `transport' quantity by a set of particles as proposed in \cite{Moore2011}. In this context, advection is applied to keep track of the velocity changes for each point, $p$ along its velocity field defined by $(u,v)$.

\begin{equation}\label{eq:ode03}
\frac{d\vec{x}_p}{dt} = u_p(t_0,t,x_0,x_p) 
\end{equation}
\begin{equation}\label{eq:ode05}
\frac{d\vec{y}_p}{dt} = v_p(t_0,t,y_0,y_p) 
\end{equation}
%
where $(x_0, y_0)$ represents the initial position of point $p$ at time $t_0$, while $(x_p, y_p)$ denotes its position at time $t_0+t$.
%
%
%
Unlike the conventional optical flow representation that captures the velocity of a pixel in two consecutive frames, the advected flow field captures the velocity of a particle in $\tau$ consecutive frames. The trace of particles over time forms a pathline. We make assumption on the initial position of $p$ as the mean velocity fields, and perform cubic interpolation of the neighboring flow field to compute the robust velocity of particles.

We adopted the Jacobian method as in~\cite{haller00} to measure the separation between each pathline which are seeded spatially close to a point, $p$, within a time instance, $\tau$. The Jacobian is computed by the partial derivatives of $d\vec{x}_p$ and $d\vec{y}_p$, where:

\begin{equation}\label{eq:jacobian01}
\nabla{F^t(p)} = \begin{bmatrix}
\frac {\partial{d\vec{x}_p}}{\partial{x_p}} & \frac {\partial{d\vec{x_p}}}{\partial{y_p}} \\
\frac {\partial{d\vec{y}_p}}{\partial{x_p}} & \frac {\partial{d\vec{y_p}}}{\partial{y_p}}
\end{bmatrix} 
\end{equation}

According to the theory of linear stability analysis in \cite{SeydelBook1994}, the square root of the largest eigenvalue, $\lambda^t(p)$ of $\trans{F^t(p)}F^t(p)$ indicates the maximum offset or displacement if the particle's seeding location is shifted by one unit as it satisfies the condition that $ln \lambda^t(p) > 0$. In the context of this study, a large eigenvalue indicates that the query point is unstable, and vice versa for a small eigenvalue. Since we are only interested in regions that have interesting motion dynamics, based on the eigenvalue, we can compute the stability of a point  using Eq. \ref{eq:stability01}. In practice, $\tau$ should depend on the rate of change of the flow field, with a higher rate of change of flow field resulting in smaller time scales and vice versa. In our experiments, we fixed $\tau=50$ frames at 25fps.

\begin{equation}\label{eq:stability01}
\phi^t = \frac{1}{\mid \tau \mid} \log \sqrt{\lambda^t(p)}
\end{equation}

This is followed by transforming the low-level feature comprising the stability coefficient, which in this study acts as an indicator of unstable motion, into global similarity structure space. The stability map is computed by taking the difference between the stability of each point, $i$, with every other point, $j$, in the given scene: 

\begin{equation}\label{eq:stability02}
s_{i,j}^t = \phi_i^t - \phi_j^t
\end{equation}

\noindent where $s_{i,j}$ is the $(i,j)$ element in the stability map denoted by $S \in \mathbb{R}^{h \times  w}$, and $h$ and $w$ represent the height and weight of the given frame.

\subsubsection{Phase Shift Map}
In order to uncover the collective flow of the crowd, one of the simplest way is `grouping' points in the velocity field, $\overline V$, according to the phase similarity. Here, we anticipated that connecting `grouped' points with respect to the gradual changes of the velocity phase, we can uncover important motion characteristic of the crowd. 

The phase shift map is denoted by $\Theta \in \mathbb{R}^{h\times w}$. Each element $\theta^t_{i,j} \in \Theta$ is obtained as the phase difference of the  mean flow vector between points:

\begin{equation}\label{eq:dotproct} 
\theta^{t}_{i,j} = \arccos \frac{\overline V_{i}^t\cdot \overline V_{j}^t}{\left \| \overline V_{i}^t \right \|\left \| \overline V_{j}^t \right \|} 
\end{equation}

\noindent where the phase difference, $\theta^{t}_{i,j}$, between two points are measured by the shortest great-circle distance, hence $\theta^{t}_{i,j}$ is bounded by [0, $\pi$]. The rational of projecting the velocity phase to the global similarity structure is to reveal the intrinsic relationship of each point, $p$, with the other points on the same video sequence.

\subsection{Saliency Detection by Manifold Ranking}
In the following, we will explain the steps to detect the salient motion regions within the crowd scene by performing ranking on the intrinsic manifold \cite{Zhou04ranking} uncovered by the global similarity feature maps, \ie~the stability and phase shift maps. 

For each video sequence, we represent the set of data points $\mathcal{R} = \left \{ \mathbf{r}_{1}, \mathbf{r}_{2}, \dots,\mathbf{r}_{n} \right \}$, in the form of a weighted k-nearest neighbors (kNN) undirected network graph $G = \left \langle V,E \right \rangle$. Note that each data point, $\mathbf{r} = \trans{(s^t,\theta^t)}$, is an integrated feature comprising the global similarity structure representation of scaled stability and phase change, where $s^t$ and $\theta^t$ are scaled to $[0,1]$. Each vertex, $\upsilon _{i}$, in the graph represents a data point, $\mathbf{r}_{i}$. Two vertices are connected by an edge $E$ weighted by a pairwise affinity matrix, $W_{ij}$, which is defined as:

\begin{equation}\label{eq:affinityMat} 
W_{ij}= \exp\left ( \frac{-\mathrm{dist}^{2}(\mathbf{r}_{i},\mathbf{r}_{j}) }{\sigma_{i}\sigma_{j}}\right )
\end{equation}

\noindent where $i\neq j$ and $W_{ii} = 0$ to avoid self reinforcement during the manifold ranking~\cite{Zhou04ranking}. $\sigma_{i}$ and $\sigma_{j}$ are the local scaling parameters~\cite{zelnik2004self}. The selection of $\sigma_{i}$ is given as:
\begin{equation}
\sigma_{i} = \mathrm{dist}(\mathbf{r}_{i},\mathbf{r}_{k}) 
\end{equation}
where $r_{k}$ is the $k$-th neighbor of data point $r_{i}$. 
%
%
The distance metric, $\mathrm{dist}$, denotes the Euclidean distance. Given the affinity matrix, $W_{ij}$, we can then represent the connected graph, $G$, using the normalized Laplacian matrix, $L= D^{-\frac{1}{2}}WD^{-\frac{1}{2}}$, where $D$ is the diagonal matrix with $D_{ii}=\sum _{j}W_{ij}$.  

We assume the typical and uninteresting motions dominate a scene. Thus, selecting a random set of $m$ `query' points, $\mathcal{Q} = \left \{\mathbf{q}_{1}, \mathbf{q}_{2}, \dots , \mathbf{q}_{m}\right \}$ can well capture the dominant crowd behavior of the scene\footnote{The selection of those random points can be repeated to generate more queries, accordingly. In this study, we set $m=100$. Evaluation with varying query points generated consistent rank score.}.  
By performing ranking, we can detect extrema as data points with the highest and lowest rank scores, deviating from the query points.
%
%
Such extrema suggest interesting regions caused by crowding, local irregular motion and sources and sinks.

To detect the extrema, we label each query successively with a positive label +1. Its label is then propagated to all other unlabeled instances, $\left\{\mathbf{r}_{i}\right\}$, of which their initial labels are assigned as $0$.
%
More precisely, we compute a rank score vector for each query $\mathbf{q}_i$, individually, denoted as $\textbf{c}_{i} = \trans{(c_{i}^{1},...,c_{i}^{n})}$, via the Laplacian graph, $L$, using the close form equation:
\begin{equation}
\textbf{c}_{i} = (I-\alpha L)^{-1}y
\end{equation}
where $I$ is an identity matrix and $\alpha$ is a scaling parameter in the range of $[0,1]$. 
The vector $y$ is the initial label assignment of data points, which is given as $y = \trans{(y_{1},...,y_{n})}$, in which $y_i = + 1$ if $\mathbf{r}_i =\mathbf{q}_i$, and $y_i = 0$ otherwise. Note that $\mathbf{q}_j$ where $j\neq i$ has initial label assigned as 0 too.
We repeat the same ranking process for all query points $\mathcal{Q}$.
The final rank score vector, $\textbf{C}$, is the average of $m$ rank score vectors, \ie~$\textbf{C}  = \frac{1}{m}\sum_{i=1}^{m} \textbf{c}_{i}$. 
Extrema are data points with the highest and the lowest rank scores in $\textbf{C}$.



\section{Experiments}

We used the benchmark datasets obtained from~\cite{Rodriguez11,ali2007lagrangian,Loy12,solmaz12} to evaluate the proposed framework. The sequences are diverse, representing dense crowd in the public spaces in various scenarios such as pilgrimage, station, marathon, rallies and stadium. In addition, the sequences have different field of views, resolutions, and exhibit a multitude of motion behaviors that cover both the obvious and subtle instabilities. 

\subsection{Qualitative Analysis}

\subsubsection{Instability Detection} 
\begin{figure}[t]
\centering
\subfigure[Original image]{
    \includegraphics[height=0.28\linewidth, width=0.45\linewidth]{Results/FW1.jpg} 
    \label{fig:subfigintro1e}
}
\subfigure[Our method]{
    \includegraphics[height=0.28\linewidth, width=0.45\linewidth]{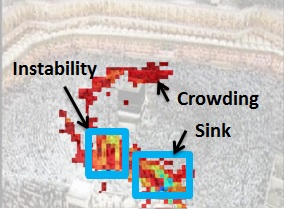} 
    \label{fig:subfigintro2e}
}
\subfigure[Loy~\etal~\cite{Loy12} ]{
    \includegraphics[height=0.28\linewidth, width=0.45\linewidth]{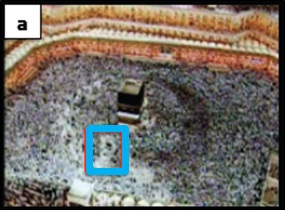} 
    \label{fig:subfigintro3e}
}
\subfigure[Ali~\etal~\cite{ali2007lagrangian}]{
    \includegraphics[height=0.28\linewidth, width=0.45\linewidth]{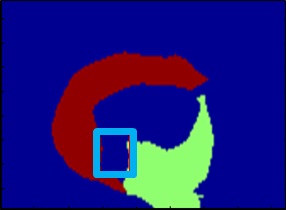} 
    \label{fig:subfigintro4e}
}
\label{fig:figurese}
\caption{Comparisons on the corrupted pilgrimage sequence, where synthetic noise was added to simulate unstable motion. Best viewed in color.}
\label{fig:mecca02}
\end{figure}

\begin{figure}[ht]
\centering
\subfigure[Original image]{
    \includegraphics[height=0.28\linewidth, width=0.45\linewidth]{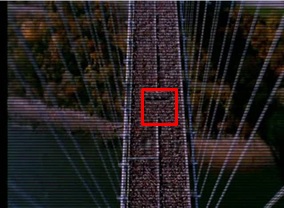} 
    \label{fig:subfigintro1b}
}
\subfigure[Our method]{
    \includegraphics[height=0.28\linewidth, width=0.45\linewidth]{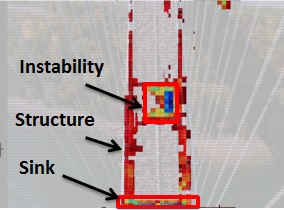} 
    \label{fig:subfigintro2b}
}
\subfigure[Loy et al. \cite{Loy12} ]{
    \includegraphics[height=0.28\linewidth, width=0.45\linewidth]{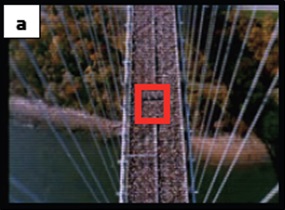} 
    \label{fig:subfigintro3b}
}
\subfigure[Ali et al. \cite{ali2007lagrangian}]{
    \includegraphics[height=0.28\linewidth, width=0.45\linewidth]{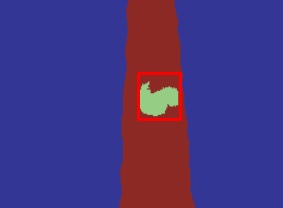} 
    \label{fig:subfigintro4b}
}
\label{fig:figuresb}
\caption{Comparisons on the corrupted marathon sequence, where synthetic noise was added to simulate unstable motion. Best viewed in color.}  
\label{fig:marathon02} 
\end{figure}

A set of two sequences comprising a pilgrimage and marathon scenes were used to test the capability of the proposed system in detecting instability. Following the studies~\cite{ali2007lagrangian,Loy12}, we introduced synthetic noise into the 2 sequences to simulate the unstable region as enclosed in the blue bounding box shown in Fig.~\ref{fig:mecca02} and the red box in Fig.~\ref{fig:marathon02}, respectively. 
We observe that all three methods (\cite{Loy12,ali2007lagrangian} and ours) are able to identify the unstable region, as shown in Fig. \ref{fig:mecca02}-\ref{fig:marathon02}. However, in addition to the synthetic noise, our proposed method is able to identify other regions that exhibit unique motion dynamics as highlighted by the colored regions. After scrutinizing our results, we notice that these areas correspond to the exit and turning point around the Kaaba in Fig.~\ref{fig:mecca02}, where there is potential slowdown in the pace of individuals, thus resulting in salient motion dynamics within these regions. Similarly, the proposed method is able to detect the sink region in the marathon sequence in Fig. \ref{fig:marathon02}, where the crowd exit from the field of view. The results demonstrate the effectiveness of the global similarity structure in capturing the intrinsic structure of the crowd motion.


\begin{figure}[t]
\centering
\subfigure[Original image]{
    \includegraphics[height=0.28\linewidth, width=0.45\linewidth]{Results/FW1.jpg} 
    \label{fig:subfigintro1}
}
\subfigure[Our method]{
    \includegraphics[height=0.28\linewidth, width=0.45\linewidth]{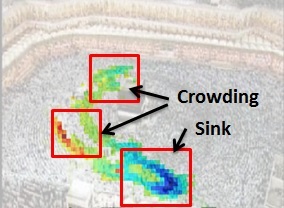} 
    \label{fig:subfigintro2}
}
\subfigure[Loy~\etal~\cite{Loy12} ]{
    \includegraphics[height=0.28\linewidth, width=0.45\linewidth]{Results/FW1.jpg} 
    \label{fig:subfigintro3}
}
\subfigure[Ali~\etal~\cite{ali2007lagrangian}]{
    \includegraphics[height=0.28\linewidth, width=0.45\linewidth]{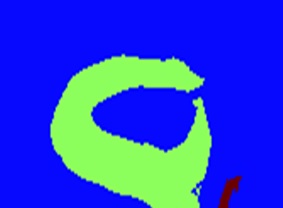} 
    \label{fig:subfigintro4}
}
\label{fig:figures}
\caption{Comparisons on the original pilgrimage sequence (without synthetic noise). Best viewed in color.}
\label{fig:mecca01}  
\end{figure}

To further evaluate the robustness of the proposed method in dealing with inconsistent and subtle crowd motion, we tested the three methods again on the original sequences of pilgrimage and marathon, without any synthetic noise. 
The results in Fig. \ref{fig:mecca01} show that~\cite{Loy12,ali2007lagrangian} do not have any detection for these sequences. In contrast, our method is capable of detecting the sink region, as well as the potential overcrowding regions along the bridge's edge. Note that the results herein are consistent with the sequences with synthetic noise since our method detect similar interesting regions. 
The results, again, show that subtle motion can be more effectively discovered by employing the global similarity structure of the crowd motion rather than using the low-level flow field~\cite{Loy12,ali2007lagrangian}
%

\begin{figure}[t]
\centering
\subfigure[Original image]{
    \includegraphics[height=0.28\linewidth, width=0.45\linewidth]{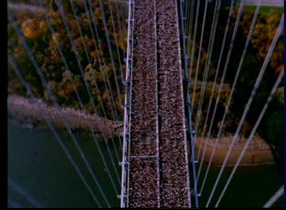} 
    \label{fig:subfigintro1c}
}
\subfigure[Our method]{
    \includegraphics[height=0.28\linewidth, width=0.45\linewidth]{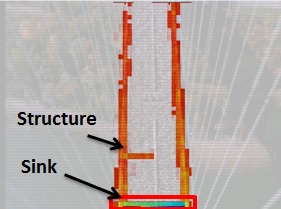} 
    \label{fig:subfigintro2c}
}
\subfigure[Loy~\etal~\cite{Loy12} ]{
    \includegraphics[height=0.28\linewidth, width=0.45\linewidth]{Results/fig6_ori.jpg} 
    \label{fig:subfigintro3c}
}
\subfigure[Ali~\etal~\cite{ali2007lagrangian}]{
    \includegraphics[height=0.28\linewidth, width=0.45\linewidth]{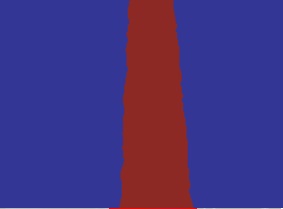} 
    \label{fig:subfigintro4c}
}
\label{fig:figuresc}
\caption{Comparisons on the original marathon sequence (without synthetic noise). Best viewed in color.}
\label{fig:marathon01} 
\end{figure}

\subsubsection{Local Irregular Motion Detection} 
\begin{figure}[t]
\centering
\subfigure[Original image]{
    \includegraphics[height=0.28\linewidth, width=0.45\linewidth]{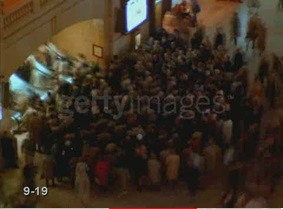} 
    \label{fig:subfigintro1a}
}
\subfigure[Solmaz~\etal~\cite{solmaz12}]{
    \includegraphics[height=0.28\linewidth, width=0.45\linewidth]{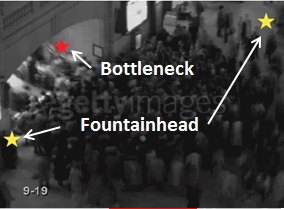} 
    \label{fig:subfigintro2a}
}
\subfigure[Our method]{
    \includegraphics[height=0.28\linewidth, width=0.95\linewidth]{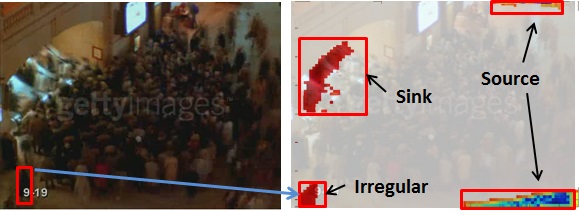} 
    \label{fig:subfigintro3a}
}

\label{fig:figuresa}
\caption{Comparison with the state-of-the-art method~\cite{solmaz12} on the station sequence. Best viewed in color.}
\label{fig:station} 
\end{figure}

Another comparison is performed between our work and Solmaz~\etal~\cite{solmaz12} using the sequence obtained from an underground station as depicted in Fig. \ref{fig:station}. 
This sequence contains obvious source and sink regions, which are detected as bottleneck and fountainhead in~\cite{solmaz12}. The results demonstrate that our method is able to detect similar regions as in~\cite{solmaz12}, with the addition of another source region at the bottom right of the scene, which is not detected by~\cite{solmaz12}. In addition, our method detected the irregular motion of someone walking into the scene from the bottom left corner of the scene. This is not the case in~\cite{solmaz12}, where their detection does not highlight accurately the location of the triggering event. Note that while our method is able to detect salient/interesting motion dynamics, we do not characterize them into the different categories. 

\begin{figure} [t]
\begin{center}  
  \includegraphics[height=0.55\linewidth, width=0.95\linewidth]{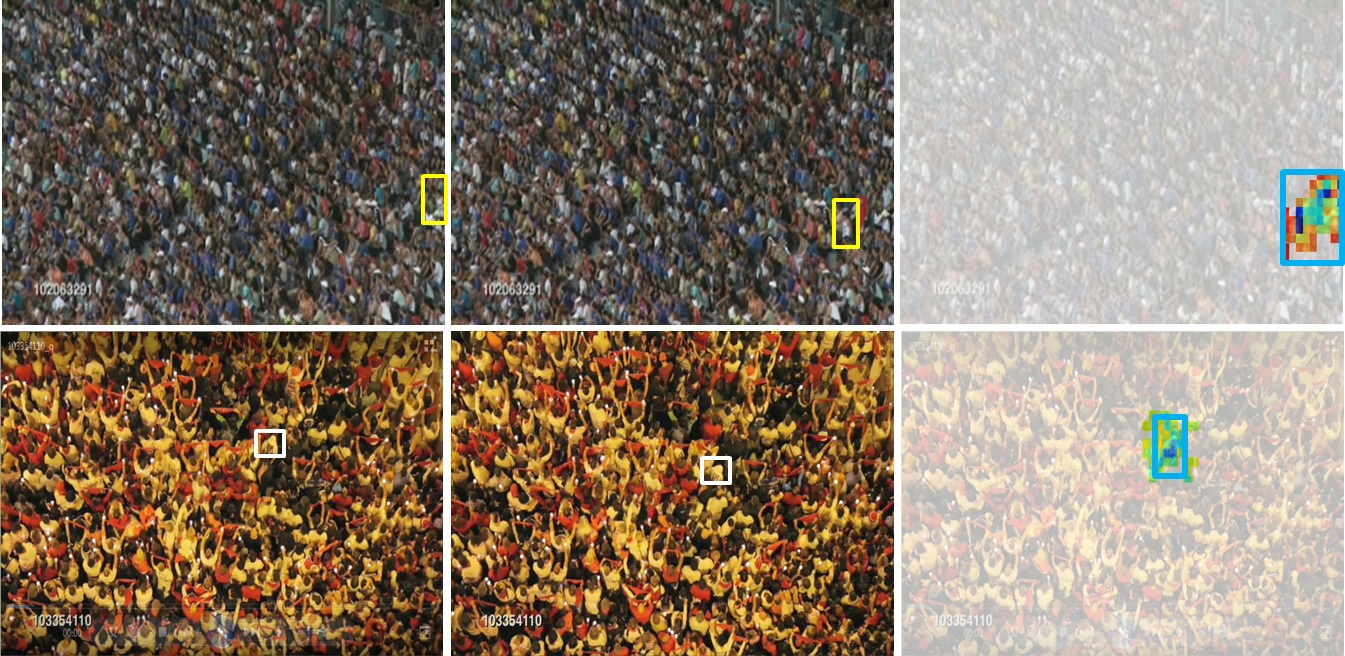}
\caption{Example detections on local irregular motion. Our output is highlighted in the blue bounding box on the right column. First row: Our method detect an individual walking across the scene, while the rest of the crowd is seated. Second row: Our method detect an individual maneuvering through an extremely crowded scene.  Best viewed in color.} \label{fig:localflow} 
\end{center}
\end{figure}

We further tested our method on sequences with local irregular motion caused by individuals moving against the dominant crowd flow such as that shown in Fig. \ref{fig:localflow}. This scenario is to mimic the Boston Marathon Person Finder page launched by Google, which aims to identify individuals that seem suspicious. Through the proposed global similarity structure of the crowd motion, our method detects such anomaly consistently and effectively, as illustrated in Fig. \ref{fig:localflow}.

\subsection{Quantitative Analysis}

We compared our detections against manually labeled interesting regions from all the sequences obtained from the public datasets. Most of the related studies ~\cite{Loy12,ali2007lagrangian}, merely provide qualitative results and the implementations are not shared publicly; leading to difficulties in performing a comprehensive evaluation quantitatively. 
We determined the regions with interesting motion dynamics as per video basis and we employed the \emph {F-measure} according to the score measurement of the well-known PASCAL challenge \cite{Everingham10}. That is, if the detected region overlaps the ground truth region by more than 50\%, then the detection is considered as the correct salient region. 

For clarity, we present our detection results according to different interesting motion categories, \ie~crowding, sources and sinks and local irregular motion, as shown in Table~\ref{table:result}. 
In general, the proposed method performs exceptionally well with only several false detections that are due to ambiguous local motion, \eg~random hand waving motion in a crowded scene. Our method fail in scenarios where the stability and phase features are derived from inaccurate flow field due to strong illumination. Specifically, the proposed ranking algorithm produce erroneous connected graphs, leading to mis-detections.  

\section{Conclusion}
We have demonstrated that the transformation of the low-level flow field descriptors, stability and phase changes, into the global similarity structure, is an effective indicator for salient motion dynamics and irregularities in the crowded scenes. 
In particular, experimental results have shown that the method is effective in detecting sources and sinks, crowding, and local irregular motions from various surveillance scenarios.
Importantly, accurate detection is achieved in the crowded scenes without tracking, prior information or model learning. 
Though the manifold projection is capable of discovering intrinsic structure of the motion dynamics, the basis of our manifold is optical flow. Thus, it is limited by the known drawbacks of optical flow estimation. Future investigation includes identifying low-level features that are more robust towards characterising motion in extremely crowded scenes.

\begin{table}[t]
\caption{Summary of the crowd saliency detection results.}
\vspace{-0.5cm}
\begin{center}
\resizebox{8.5cm}{!}{
    \begin{tabular}{|l||c|c|c|c|}
    \hline
    Motion Category & Total \# & \# of & \# of Missed & \# of False \\
     & of Labelled Region & Detection & Detection & Detection \\
    \hline
    \hline
    Crowding & 13  & 12  & 1 & 0 \\
    \hline
    Sources \&
    Sinks    & 19  & 14  & 5  & 0               \\
    \hline
    Local Irregularity
     & 43 & 47 & 2 & 6               \\
    \hline
    \end{tabular}}
    \label{table:result}
    \end{center}
    \vspace{-0.25cm}
\end{table}

\section*{Acknowledgement}
This research is supported by the Fundamental Research Grant Scheme (FRGS) MoE Grant FP027-2013A, H-00000-60010-E13110 from the Ministry of Education Malaysia.

{\small
\bibliographystyle{IEEEtran}
\bibliography{ReferenceICPR}
}

\end{document}